\documentclass[10pt,twocolumn,letterpaper]{article}

\usepackage{iccv}
\usepackage{times}
\usepackage{epsfig}
\usepackage{graphicx}
\usepackage{amsmath}
\usepackage{amssymb}
\usepackage{tabularx}
\usepackage{algorithm}
\usepackage{algpseudocode} 

\usepackage[breaklinks=true,bookmarks=false]{hyperref}

\iccvfinalcopy 

\def\iccvPaperID{9508} 

\ificcvfinal\fi

\begin{document}

\title{Attention Map-Guided Two-stage Anomaly Detection using Hard Augmentation}


\author{
Jou Won Song, Kyeongbo Kong, Ye In Park, Suk-Ju Kang\\
Department of Electronic Engineering, Sogang University , Seoul, South Korea\\
{\tt\small sjkang@sogang.ac.kr}}

\maketitle
\ificcvfinal\thispagestyle{empty}\fi

\begin{abstract}
Anomaly detection is a task that recognizes whether an input sample is included in the distribution of a target normal class or an anomaly class. Conventional generative adversarial network (GAN)-based methods utilize an entire image including foreground and background as an input. However, in these methods, a useless region unrelated to the normal class (e.g., unrelated background) is learned as normal class distribution, thereby leading to false detection. To alleviate this problem, this paper proposes a novel two-stage network consisting of an attention network and an anomaly detection GAN (ADGAN). The attention network generates an attention map that can indicate the region representing the normal class distribution. To generate an accurate attention map, we propose the attention loss and the adversarial anomaly loss based on synthetic anomaly samples generated from hard augmentation. By applying the attention map to an image feature map, ADGAN learns the normal class distribution from which the useless region is removed, and it is possible to greatly reduce the problem difficulty of the anomaly detection task. Additionally, the estimated attention map can be used for anomaly segmentation because it can distinguish between normal and anomaly regions. As a result, the proposed method outperforms the state-of-the-art anomaly detection and anomaly segmentation methods for widely used datasets.

\end{abstract}
\begin{figure}[t]
\begin{center}
\includegraphics[width=0.9\linewidth]{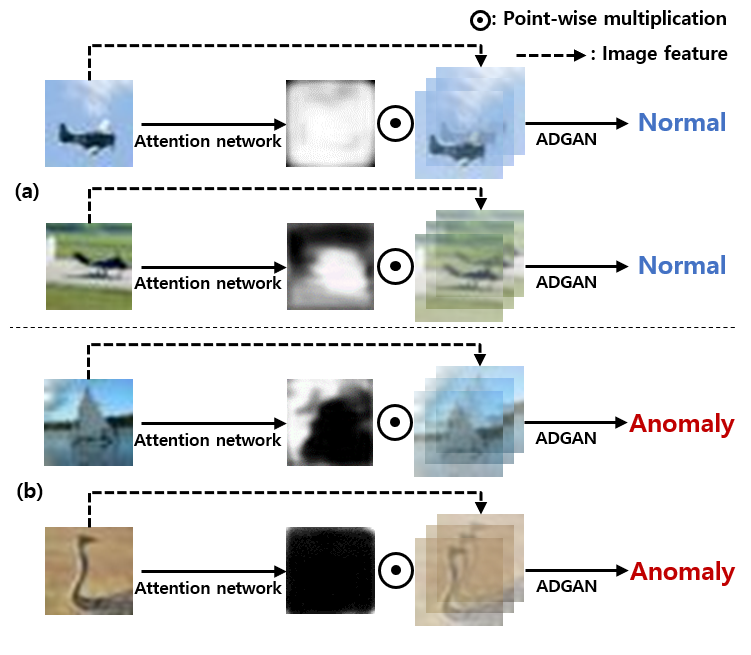} 
\end{center}
   \caption{The anomaly detection process of the proposed method. The normal class is the airplane of the CIFAR10. (a) The attention map focuses on the sky or objects related to the airplane. In this case, ADGAN can detect the anomaly class more sensitively by learning the image feature from which the useless region has been removed. (b) In the anomaly class, the attention map removes a useless region except for the sky, which is the normal region. Since the useless features multiplied by the attention map are not learned in the ADGAN, the proposed detection method can discriminate the anomaly class.}
\label{fig1}
\vspace{-0.4cm}
\end{figure}
\section{Introduction}
Anomaly detection is a process that determines whether a given input sample is included in a trained class. It is applied in various fields, such as industrial inspection for segmenting defective product parts \cite{mvtec}, unbalanced learning \cite{re4}, and intruder detection \cite{re1,re2,re3}. Typical anomaly detection tasks work on datasets with a large number of normal class images and a small number of undefined anomaly class images. Generally, conventional classification methods are ineffective for these datasets because of the class imbalance problem. Accordingly, the standard approach of existing methods is to learn a representation of the normal class distribution using a variational auto-encoder (VAE) \cite{re8,re9,re12,re10} or generative adversarial network (GAN) \cite{re7,re11,anog,re14}. When the anomaly class image is used as an input to these networks, anomaly detection can be performed due to a large difference from the normal class output is observed.

Generally, existing methods, which exploit the VAE (or GAN)-based network, use the whole region of an input image. However, these methods encounter the problem that a useless region unrelated to the normal class is learned as the normal class distribution. Fig. \ref{fig1} shows an example of this problem and the main concept of our approach when the normal class is airplane class. In common airplane images, the sky image is the most common background, whereas the grass is normally unrelated to airplanes. If a useless region, such as the grass background, is learned as a normal class distribution, an anomaly class may be incorrectly detected as a normal class. The proposed attention map, on the other hand, focuses on the region related to the normal class. The anomaly detection GAN (ADGAN) is only trained with the region for the normal class after removing the useless region. Hence, ADGAN stably learns the normal class distribution. The detection process for the anomaly class is shown in Fig. \ref{fig1} (b). Anomaly images (boat, bird) were not trained in the attention network. Therefore, the attention map removes useless regions except for the sky, which belongs to the normal region. The feature multiplied by this attention map extremely differs from that of the normal class distribution, and ADGAN can easily discriminate this image as that of an anomaly class.

This paper proposes a novel two-stage structure consisting of an attention network that generates an attention map to remove the useless region that is unrelated to the normal class and the ADGAN that learns the distribution from which the useless region is removed. The goal of the attention network is to generate an attention map that represents the normal class region within a given image for anomaly segmentation. To accomplish this, adversarial anomaly loss, attention loss, and hard augmentation, which generates data shifted away from the input data distribution, are employed in the training stage. 

As confirmed in \cite{att} and \cite{eatt}, a semi-supervised learning using a small amount of anomaly data improves the performance of the deep learning model. Therefore, the proposed method uses hard augmented anomaly data extracted from normal data instead of real anomaly data to improves performance. In addition, the adversarial anomaly loss aids the attention map to deactivate the useless region in the image, and the attention loss induces to distinguish accurately between anomaly and normal regions at the pixel level. Then, through pixel-wise multiplication between the image feature and attention map, the useless region unrelated to the normal class is removed from the image feature. Finally, the proposed ADGAN uses this feature to train the normal class distribution after removing the useless region. The main contributions of this study are summarized as follows:
\\
\\
- We propose a novel GAN-based attention network to generate an attention map that focuses on the normal class region. We also introduce an attention loss and adversarial anomaly loss using the hard augmentation to help perform accurate anomaly segmentation. \\
- The two-stage network structure, which connects the attention network and ADGAN, is proposed to learn the feature distribution after removing a useless region. The ADGAN learns the normal class distribution from which the useless region is removed, and hence, it can detect the anomaly class more sensitively than existing methods.\\ 
- We confirmed that applying hard augmentation improves the performance even in the GAN-based anomaly detection, thereby improving the performance of the proposed model.\\
- The proposed method outperforms the state-of-the-art methods on several datasets in terms of the classification accuracy and the average area under a receiver operating characteristic curve (AUROC).
\begin{figure*}
\begin{center}
\includegraphics[width=0.9\linewidth]{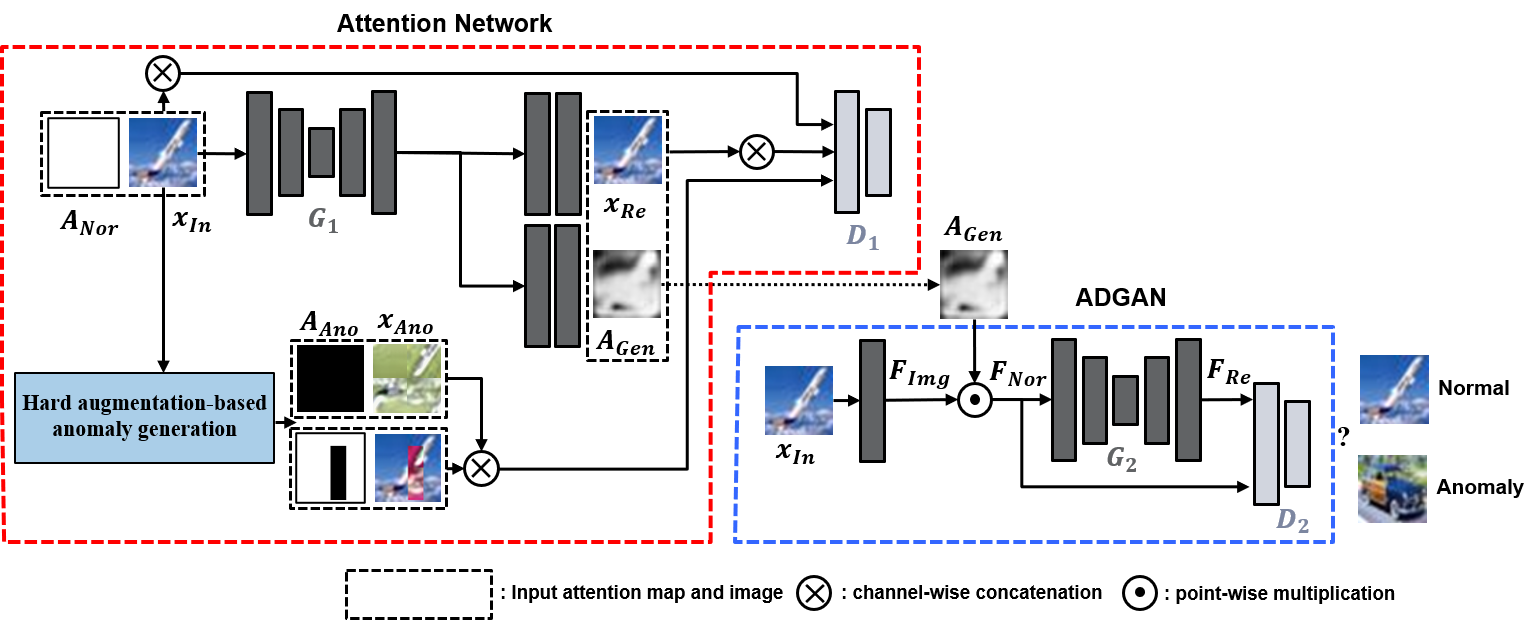} 
\end{center}
   \caption{Overall network structure of the proposed method. The network consists of two sub-networks: the attention network and ADGAN.}
   \vspace{-0.2cm}
\label{fig2}

\end{figure*}

\begin{figure}[t]
\begin{center}
\includegraphics[width=0.8\linewidth]{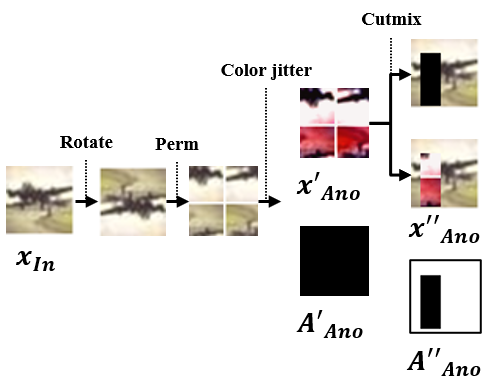} 
\end{center}
   \caption{Our hard augmentation-based anomaly generation. Two types of synthetic anomaly data $x_{Ano}$ is generated by the several hard augmentations. $x^{'}_{Ano}$ is generated by applying a rotation, perm, and color jitter for each step, and $x^{''}_{Ano}$ is generated in a similar method to Cutmix \cite{cutmix} at the last step.}
      \vspace{-0.4cm}
\label{fig3}

\end{figure}

\section{Related Works}

Anomaly detection is a research topic that has received the considerable attention in the past. In existing methods, this task is also known as one-class classification \cite{ocsvm,ocgan} or out-of-distribution detection \cite{re6}. Anomaly detection is usually performed via unsupervised methods using the generation model for learning the distribution of a certain class. In these methods, GAN \cite{gan} (or VAE \cite{vae}) learns the distribution of a certain class and uses the difference between a reconstructed image and an input for anomaly detection. An et al. \cite{re8} proposed a simple reconstruction error-based anomaly detection with the VAE. Chen et al. \cite{re10} introduced autoencoder ensembles for unsupervised anomaly detection. Sakurada et al. \cite{re12} proposed using the mean square error to perform anomaly detection. Additionally, several methods for performing anomaly detection using the property of GAN \cite{gan} that learns the input class distribution have been also studied. In \cite{alocc}, GAN was trained to add noise to the normal class images to eliminate noise. These refined images are generated closer to the distribution of the training class.  

In addition, deep learning-based anomaly segmentation methods focus on generative models such as GAN \cite{anog} and VAE \cite{mvtec}. However, these approaches usually may have a high reconstruction performance for anomaly class images because anomaly image regions may exist in normal images used for learning. In \cite{stu}, student networks were trained to mimic a pretrained teacher dividing an image into patches. However, the method of using an image patch often has a problem that performance varies greatly depending on the patch size. In recent works, an activation map that visualizes the region of interest (ROI) through grad-cam \cite{grad} was applied to anomaly detection. Kimura et al. \cite{att} focused on realistic dataset settings with complex noise and considerably few training images. The activation map was generated using grad-cam to focus only on the reconstruction loss of the ROI. Venkataramanan et al. \cite{eatt} improved the detection performance using an activation map in the training process. However, the disadvantage of these methods is that the noise unrelated to the normal class is still used in training, and an incorrectly predicted activation map cannot be used. The proposed method directly extracts and supervises the attention map. Our method is different from the conventional method which uses grad-cam \cite{grad} to indirectly extract the activation map. Therefore, the proposed method distinguishes between anomaly and normal regions more accurately than previous methods.

\section{Proposed Method}
The proposed network consists of two parts: the attention network that generates an attention map and ADGAN that learns a normal class distribution after removing the useless region by the attention network. The details are explained in the following sections.

\subsection{Overall Proposed Model}
The overall structure of the proposed network is shown in Fig. \ref{fig2}. The attention network consists of a generator and a discriminator. Unlike the existing GAN structure, the generator of the attention network generates an reconstructed image $x_{Re}$ and an generated attention map $A_{Gen}$. The discriminator of the attention network also determines whether an input image and an attention map are real or fake. Furthermore, anomaly data $x_{Ano}$ generated by the hard augmentation \cite{csi}, which generates data that is shifted away from the input data distribution, is used to support the attention network estimates normal data distribution. Then, the attention map activates the normal region and deactivates the useless region in an input image. Therefore, the attention map $A_{Nor}$ from normal data is fully activated while the attention map $A_{Ano}$ from the anomaly data $x_{Ano}$ is deactivated. The attention network is trained through adversarial training of GAN. The generator aims to generate an attention map that activates the normal region of an input image while deactivating the useless region. The attention network yields the feature map $F_{Nor}$ with the useless region removed, thereby maintaining the normal feature intact. In this stage, the feature map $F_{Nor}$ is acquired by multiplying the attention map and the image feature $F_{Img}$ extracted from CNN. Finally, ADGAN trains only the normal region distribution using $F_{Nor}$ to perform anomaly detection.

\subsection{Attention Network}
\subsubsection{Hard Augmentation-based Anomaly Generation}
The proposed method uses several hard augmentations to obtain the synthetic anomaly data different from the normal data distribution. It helps to estimate the exact normal region in the input image. We performed an experiment in section 4.2.1 using DCGAN \cite{dcgan} to apply the anomaly data generation. Through this experiment, we observed that using several hard augmentations in the training process helps the training model to estimate the normal data distribution. According to the results of this experiment, our anomaly generation used three types of hard augmentation: rotation, color jitter, and perm. Fig. \ref{fig3} shows the process of generating synthetic anomaly data in the proposed anomaly generation. The anomaly generation generates two types of anomaly data $x_{Ano}$. The first synthesized anomaly data $x^{'}_{Ano}$ is generated using three augmentations previously introduced. Similar to CutMix \cite{cutmix}, a random region is selected to generate the second anomaly data $x^{''}_{Ano}$. The selected region is then replaced with the zero value or the corresponding region of $x^{'}_{Ano}$, which have the same chance. In addition, since we know which region of the generated $x_{Ano}$ is a useless region, the proposed anomaly generation obtains $A^{'}_{Ano}$ and $A^{''}_{Ano}$ corresponding to $x^{'}_{Ano}$ and $x^{'}_{Ano}$. Effectiveness of synthetic anomaly data is verified in Section 4.2.1.

\subsubsection{GAN-based Attention Network}
The meaningful region in the attention map is obtained through the training process of the attention network. GAN, which is generally used in anomaly detection, can easily detect the anomaly class because it learns the normal class distribution. To reflect this property on the attention map, we design the generator to create both the attention map and the reconstructed image. Fig. \ref{fig2} shows the structure of the designed generator $G_1$. The attention map $A_{Gen}$ has the same size as an input image $x_{In}$ and outputs a value in the range of [0, 1] depending on the importance of the pixel in an input image. The reconstructed image and the attention map are combined via the channel-wise concatenation, Then, it is used as an input to the discriminator. The loss function for learning the discriminator is as follows:
\begin{align}
  L_{Adv} = \underset{G}{min} \underset{D}{max}\{\mathbb{E}\;[\log(1-D(concat(x_{Re},A_{Gen})))] \nonumber \\   +\mathbb{E}\;[\log(D(concat(x_{In},A_{Nor})))]\},
  \label{equ:dt}
\end{align}
where $x$, $x_{Re}$, $D$, $G$, and $concat$ indicate an input image, a reconstructed image, a discriminator, a generator, and a concatenation operation, respectively. Therefore, the discriminator of the attention network verifies whether the reconstructed image is real or fake and whether the attention map focuses on the normal region of the image. However, if the existing GAN loss is applied only using the attention map of the normal image during the training process, the useless region can be falsely activated because the attention map activated by the proposed model can be determined as the normal distribution. To solve this problem, we train a discriminator with the adversarial anomaly loss $L_{Adv}^{ano}$ using $x_{Ano}$ and $A_{Ano}$ generated in the anomaly generation presented in the previous section. The adversarial anomaly loss equation is as follows:
\begin{align}
    L_{Adv}^{ano} = \underset{G}{min}\underset{D}{max}\{\mathbb{E}\;[1-\log(D(concat(x_{Re},A_{Gen}^{ano})))] \nonumber \\  +\mathbb{E}\;[\log(D(concat(x_{Ano},A_{Ano})))]\},
  \label{equ:dt}
\end{align}
where $A_{Gen}^{ano}$ indicates the attention map generated from the anomaly data. Therefore, the discriminator additionally learns an attention map that deactivates the region of $x_{Ano}$ , which differs from the normal data distribution. In addition, to better represent the normal class distribution, the adversarial variational autoencoder structure used in a previous study \cite{eatt} is used as the generator of GAN. The loss function for learning the generator are as follows:
\begin{align}
 L_{G} = L_{R}(x_{In}, x_{Re}) + KL(q_{\phi}(z|x)\parallel p_{\theta}(z|x)),
  \label{equ:dt}
\end{align}
where $L_{R}$ is negative log likelihood of the given input and $KL$ is the Kullback-Leibler (KL) divergence.

\subsubsection{Attention Loss}
By applying the adversarial anomaly loss, the proposed method can determine the location of the useless region of a given input image. However, it is difficult to determine the accurate useless region in the pixel level with the adversarial anomaly loss alone. Therefore, we propose the attention loss to accurately classify the useless region. The equation for the attention loss ($L_{Att}$) function is as follows:
\begin{equation}
  L_{Att} = \mathbb{E}\parallel A_{Gen}^{nor}-\,A_{Nor} \parallel ^{2} + E\parallel A_{Gen}^{ano}-\,A_{Ano} \parallel ^{2},
  \label{equ:dt}
\end{equation}
where $A_{Gen}^{nor}$ indicates the attention map generated from the normal data. As shown in the equation, the attention loss function activates and deactivates the normal and anomaly regions using the attention map, respectively. The total loss $L_{total}$ including the attention loss is as follows:
\begin{equation}
  L_{total} = L_{Adv}^{ano} + L_{Adv} + L_{G} + L_{Att}.
  \label{equ:dt}
  \vspace{-0.005cm}
\end{equation}

\subsection{Anomaly Detection GAN}

As shown in Fig. \ref{fig2}, the attention map of attention network $A_{Gen}$ is used as an input of ADGAN. The anomaly detection is performed through the ADGAN that learns distribution of $F_{Nor}$. The feature map $F_{Nor}$ is obtained by the following multiplication:
\begin{equation}
  F_{Nor} = F_{Img}\circ\,A_{Gen},
  \label{equ:dt}
\end{equation}
where $\circ$ indicates point-wise multiplication and $F_{Img}$ indicates the image feature obtained through two CNN layers. Different from the existing GAN, which learns using images, the ADGAN is trained using the image feature. The attention map has a high value in a normal class region and a low value in an anomaly class region. However, since the image pixel intensity does not represent the absolute importance, directly multiplying the image with the attention map may ignore the dark regions where the pixel intensity is zero. To overcome this problem, the image feature map is used instead of the image itself. Important information of the image feature map can be preserved because the feature map has high values in regions that the model deems important. As the feature multiplied with the attention map is used, only the distribution of the regions more related to the normal class is learned by ADGAN. Moreover, the generated anomaly class attention map is different from the normal class. Consequently, the ADGAN becomes more sensitive to the anomaly class distribution.

\begin{figure}
\begin{center}
\includegraphics[width=0.9\linewidth]{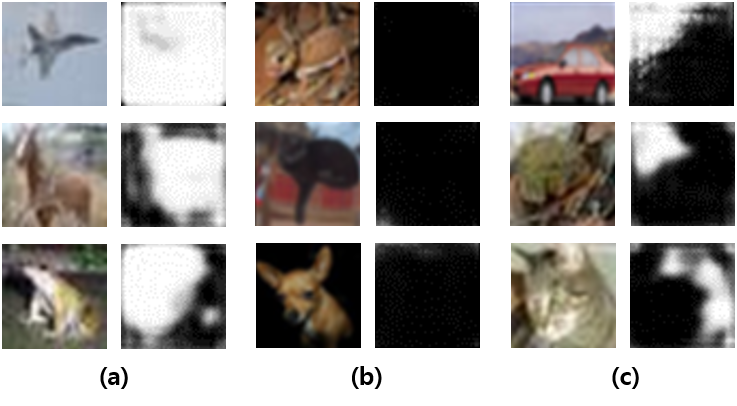} 
\end{center}
   \caption{Qualitative results on the CIFAR10 dataset. (a) normal input images and attention maps. (b), (c) anomaly input images and attention maps. The proposed method removed the region of the anomaly class. Therefore, the attention map generated by the attention network differs from the attention map of a normal image.}
     \vspace{-0.4cm}
\label{fig4}
\end{figure}

\section{Experiments}
To verify the performance of the proposed method, several evaluations were performed on CIFAR10 \cite{cifar} and MVTec Anomaly Detection (MVTec AD) datasets \cite{mvtec}. The proposed method used the CIFAR10 dataset to evaluate the anomaly detection performance. The experiment of anomaly detection was performed under the same condition as \cite{ocgan} (One class is considered normal and other classes are considered anomaly.).

In the first experiment, we observed several hard augmentations using DCGAN \cite{dcgan} to generate the anomaly data, which helps to estimate the normal data distribution. Then, we used hard augmentations to learn the proposed model and compared it with the existing methods. As a measure of anomaly detection performance, the AUROC was evaluated. We also used the MVTec AD dataset to evaluate the proposed method in terms of anomaly segmentation and anomaly detection. Following \cite{mvtec}, classification accuracy was used as an evaluation metric for the anomaly detection, and the pixel-wise mean AUROC was used for the anomaly segmentation. Finally, an output of the ADGAN discriminator and the attention map score was used as the anomaly detection and the anomaly segmentation score of the proposed method, respectively.
\begin{figure}[t]
\begin{center}
\includegraphics[width=1\linewidth]{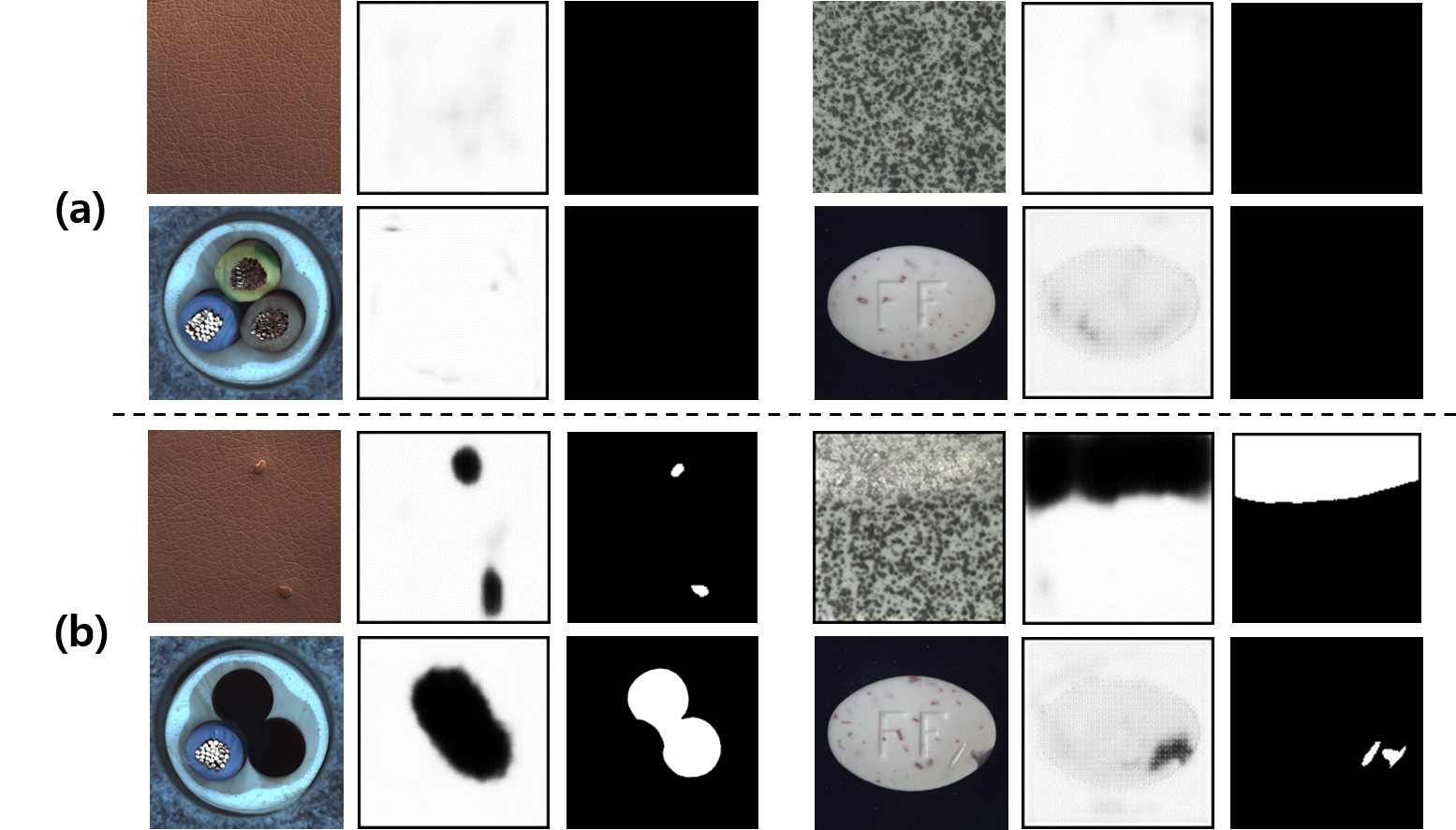} 
\end{center}
   \caption{Input images, attention maps, and ground truths on MVtec AD dataset for (a) normal class and (b) anomaly class. The attention map of anomaly class was different from the attention map of the normal class due to the defective region.}
   
\label{fig5}
\end{figure}

\begin{table}
\begin{center}
\label{table:headings}
\renewcommand{\tabcolsep}{4pt}
\makeatletter
\def\hlinewd#1{%
\noalign{\ifnum0=‘}\fi\hrule \@height #1 \futurelet
\reserved@a\@xhline}
\newcommand{\hthickline}{\hlinewd{1pt}}
\newcommand{\hthinline}{\hlinewd{.2pt}}
\makeatother
\newcolumntype{Z}{>{\centering\arraybackslash}X}
{\small
\begin{tabularx}{\linewidth}{c||Z|Z|Z|Z|Z||Z}
\hthickline
  &\multicolumn{6}{c}{$D_{C}$}\\\hline
Method &plane &car &bird &cat &deer &mean\\
\hline\noalign{\smallskip}
\hline
Base &0.601 &0.473 &0.546 &0.508 &0.624 &0.583\\\hline
Rotate &0.542 &0.499 &0.510 &0.561 &0.636 &0.612\\\hline
Perm &0.673 &0.525 &0.554 &0.587 &0.642 &0.634\\\hline
Jitter &0.582 &0.421 &0.542 &0.541 &0.627 &0.542\\\hline
$R$\&$P$ &\textcolor{blue}{0.692} &0.511 &0.532 &0.577 &0.647 &0.647\\\hline
$R$\&$P$\&$J$ &0.691 &\textcolor{blue}{0.553} &\textcolor{blue}{0.563} &\textcolor{blue}{0.592} &\textcolor{red}{0.651} &\textcolor{blue}{0.663}\\\hline
Semi &\textcolor{red}{0.720} &\textcolor{red}{0.569} &\textcolor{red}{0.591} &\textcolor{red}{0.644} &\textcolor{blue}{0.648} &\textcolor{red}{0.687}\\\hline
\end{tabularx}}
\end{center}
\caption{Performance comparison in terms of mean AUROC of 5 randomly selected classes in the CIFAR10 dataset for evaluation of anomaly detection. The mean of all ten classes is reported. Semi is a model trained with semi-supervised learning. $R$, $P$, and $J$ indicate rotation, perm, and color jitter, respectively.}
\vspace{-0.4cm}
\end{table}

\begin{table*}
\begin{center}
\label{table:headings}
\renewcommand{\tabcolsep}{4pt}
\makeatletter
\def\hlinewd#1{%
\noalign{\ifnum0=‘}\fi\hrule \@height #1 \futurelet
\reserved@a\@xhline}
\newcommand{\hthickline}{\hlinewd{1pt}}
\newcommand{\hthinline}{\hlinewd{.2pt}}
\makeatother
\newcolumntype{Z}{>{\centering\arraybackslash}X}
{\small
\begin{tabularx}{\linewidth}{c|Z|Z|Z|Z|Z|Z|Z|Z}
\hthickline
  &\multicolumn{8}{c}{$D_{C}$-Anomaly Detection (AUROC)}\\\hline
Method &AnoGAN &OCGAN &ULSLM &LSA &$\gamma$-VAE &CAVGA-D &Student &Proposed \\
\hline\noalign{\smallskip}
\hline
Mean &0.619 &0.656 &0.736 &0.641 &0.717 &0.737 &\textcolor{blue}{0.813} &\textcolor{red}{0.823}\\
\hthickline 
\end{tabularx}}
\end{center}
\caption{Performance comparison of the anomaly detection in term of mean AUROC with the proposed and existing SOTA methods on CIFAR10 ($D_{C}$) dataset. (The best result is red and the second-best result is blue.).}
\vspace{-0.0cm}
\end{table*}

\begin{table*}
\begin{center}
\label{table:headings}
\renewcommand{\tabcolsep}{0.1pt}
\makeatletter
\def\hlinewd#1{%
\noalign{\ifnum0=‘}\fi\hrule \@height #1 \futurelet
\reserved@a\@xhline}
\newcommand{\hthickline}{\hlinewd{1pt}}
\newcommand{\hthinline}{\hlinewd{.2pt}}
\makeatother
\newcolumntype{Z}{>{\centering\arraybackslash}X}
{\small
\begin{tabularx}{\linewidth}{c||Z|Z|Z|Z|Z|Z|Z|Z}
\hthickline
  &\multicolumn{7}{c}{$D_{V}$-Anomaly Segmentation (AUROC)}\\\hline
Method &AnoGAN &$AE_{SS}$ &$AE_{L2}$ &VEVAE &CAVGA-R &Superpixel &Student &Proposed \\
\hline\noalign{\smallskip}
\hline
Bottle &0.86 &\textcolor{black}{0.93} &0.86 &0.87 &0.89 &0.86 &0.85  &\textcolor{red}{0.95}\\\hline
Cable &0.78 &0.82 &0.86 &\textcolor{black}{0.90} &0.85 &\textcolor{red}{0.92} &\textcolor{black}{0.73}  &\textcolor{red}{0.92}  \\\hline
Capsule &0.84 &\textcolor{black}{0.94} &0.88 &0.74 &\textcolor{black}{0.95} &\textcolor{black}{0.93} &0.82  &\textcolor{red}{0.96} \\\hline
Carpet &0.54 &0.87 &0.59 &0.78 &\textcolor{red}{0.88} &\textcolor{red}{0.88} &0.86  &\textcolor{black}{0.86} \\\hline
Grid &0.58 &\textcolor{black}{0.94} &0.90 &0.73 &0.95 &\textcolor{red}{0.97} &0.60  &0.92  \\\hline
Hazelnut &0.87 &\textcolor{black}{0.97} &\textcolor{black}{0.95} &\textcolor{red}{0.98} &\textcolor{black}{0.96} &\textcolor{black}{0.97} &\textcolor{black}{0.91} &0.95\\\hline
Leather &0.64 &\textcolor{black}{0.78} &\textcolor{black}{0.75} &\textcolor{black}{0.95} &\textcolor{black}{0.94} &\textcolor{black}{0.86} &\textcolor{black}{0.93}  &\textcolor{red}{0.96}\\\hline
Metal Nut &0.76 &\textcolor{black}{0.89} &\textcolor{black}{0.86} &\textcolor{black}{0.94} &\textcolor{black}{0.85} &\textcolor{black}{0.92} &\textcolor{black}{0.58}  &\textcolor{red}{0.95}\\\hline
Pill &0.87 &0.91 &0.85 &0.83 &\textcolor{red}{0.94} &0.92 &0.90  &\textcolor{black}{0.91} \\\hline
Screw &0.80 &0.96 &0.96 &\textcolor{red}{0.97} &0.85 &0.96 &0.90 &0.86 \\\hline
Tile &0.50 &0.59 &0.51 &0.80 &0.80 &0.62 &0.87  &\textcolor{red}{0.93}  \\\hline
Toothbrush &0.90 &0.92 &0.93 &0.94 &0.91 &\textcolor{red}{0.96} &0.81  &0.95 \\\hline
Transistor &0.80 &0.90 &0.86 &\textcolor{red}{0.93} &0.85 &0.85 &0.85 &0.91 \\\hline
Wood &0.62 &0.73 &0.73 &0.77 &0.86 &0.80 &0.68 &\textcolor{red}{0.90} \\\hline
Zipper &0.78 &0.88 &0.77 &0.78 &\textcolor{black}{0.94} &0.90 &0.90  &\textcolor{red}{0.94} \\\hline\hline
Mean &0.74 &0.86 &0.82 &0.86 &0.89 &0.89 &0.80  &\textcolor{red}{0.92}\\\hline\hline
  &\multicolumn{7}{c}{$D_{V}$-Anomaly Detection (Classification Accuracy)}\\\hline
Mean &0.55 &0.63 &0.71 &- &0.82 &0.76 &0.84 &\textcolor{red}{0.89}\\\hline
\hthickline 
\end{tabularx}
}
\end{center}
\caption{Performance comparison of anomaly detection in term of AUROC, mean AUROC, and classification accuracy with the proposed and conventional SOTA methods on MVTec AD ($D_{V}$) dataset. (The best result is red.).}
\vspace{-0.2cm}
\end{table*}

\subsection{Implementation Details}
As shown in Fig. \ref{fig2}, the encoder of the attention network consisted the convolution layers of ResNet-18 \cite{res} pretrained from ImageNet \cite{net}. The decoder of the attention network had four transposed convolution layers that generate a feature map with the same size as an input image. In addition, two paths with three convolution layers were added to the decoder, which outputs a reconstructed image and an attention map from a feature map. The encoder of ADGAN was the same as the encoder of the attention network, and the decoder of the ADGAN consisted of four transposed convolution layers. Also, we used the discriminator of DCGAN. Finally, to generate an attention map that focuses on the normal region, the attention network is trained first and ADGAN is trained afterwards. Detailed information on training process and network architectures is described in our supplementary material.

\subsection{Experimental Results}
\subsubsection{Effectiveness of Hard Augmentation}
To verify the effectiveness of hard augmentation on the anomaly detection performance, we performed experiments to apply several hard augmentations to DCGAN. An encoder was added to the existing DCGAN structure because we followed the same procedure as defined in \cite{alocc}. First, we considered rotation and perm as introduced in \cite{csi}. Additionally, color jitter was considered to obtain various color distributions of the generated anomaly images. In semi-supervised setting, the discriminator of DCGAN is learned to discriminate the anomaly data as a fake, which is 0. Table 1 shows performances based on AUROC, and the anomaly score uses the discriminator output. ``Base'' means DCGAN learned by using only normal data. In models excluding ``Base'', the anomaly data generated by hard augmentation was used, and in ``Semi'', 128 real anomaly data was randomly selected, and DCGAN was trained.

The model trained with semi-supervised learning shows a higher performance improvement than the base model. On the other hand, the model using only one augmentation does not show as high performance improvement as ``Semi''. Especially, in the case of rotation augmentation, the flying direction is not constant for bird and airplane classes. This means that these classes have a distribution similar to the normal data distribution even after rotation augmentation is applied. Therefore, as shown in Table 1, the model to which rotation is applied shows low performance in the bird and airplane classes. However, the model using several augmentations shows higher performance than other models regardless of class types. Intuitively, using several augmentation methods generates anomaly data that is shifted from the normal data distribution, even if the property of the class is different. The results in Table 1 show that the method of using anomaly data helps to estimate the normal distribution. Therefore, by using the proposed anomaly generation, the synthetic anomaly data was generated as described in section 3.2.1 using three augmentations: rotation, color jitter, and perm.

\subsubsection{Evaluation for CIFAR10 Dataset}
To evaluate the proposed method in terms of the quantitative results, its performance was compared with other existing anomaly detection methods, which are AnoGAN \cite{anog}, OCGAN \cite{ocgan}, ULSLM \cite{ulslm}, LSA \cite{lsa}, $\gamma$-VAE \cite{gvae}, CAVGA-D \cite{eatt} and Uninformed Students (Student) \cite{stu}, using the AUROC metric described in Section 4.1. In Tables 2, the best score and the second-best score in each row were highlighted in red and blue, respectively. The proposed method achieved high performance than existing methods. 

In addition, for the CIFAR10 dataset, we observed the input image and the attention map in Fig. \ref{fig4} to verify that the attention map focuses on the normal region. Fig. \ref{fig4} (a) shows the attention map that captures the important region in an input image. Through this attention map, ADGAN could learn the normal distribution without the useless region, and effectively detect anomaly data. Figs. \ref{fig4} (b) and (c) show the anomaly input image and the attention map. If a useless region is detected in the proposed attention network, all the useless regions are removed as shown in Fig. \ref{fig4}. (b). However, the CIFAR10 dataset has a complex image distribution. Therefore, the proposed method may not be able to deactivate the corresponding region even for an anomaly class image as shown in Fig. \ref{fig4}. (c). In particular, in the first row of Fig. \ref{fig4} (c), since the normal class is an airplane, the sky region is regarded as a normal region even though it is an anomaly class image. However, since this attention map generates the feature that was not learned in ADGAN, the proposed ADGAN easily discriminated between normal class and anomaly class.

\begin{figure}[t]
\begin{center}
\includegraphics[width=1\linewidth]{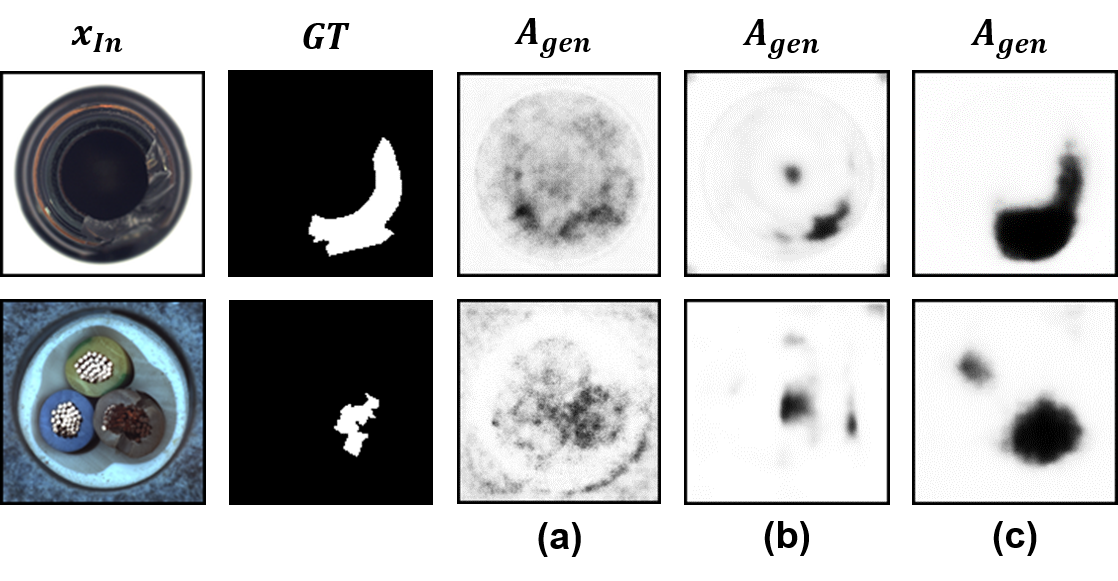}
\end{center}
   \caption{Qualitative results of the ablation study to illustrate the performance of the anomaly segmentation on the MVtec AD dataset (a) with $L_{Adv}^{ano}$, (b) with $L_{Att}$, and (c) with $L_{Adv}^{ano}$ and $L_{Att}$.}
  \vspace{-0.2cm}
\label{fig6}
\end{figure}

\subsubsection{Evaluation for MVTec AD Dataset}
We evaluated the anomaly segmentation performance between the proposed method and the existing state-of-the-art methods using the MVtec AD dataset. Existing methods were AnoGAN \cite{anog}, Autoencoder-SSIM ($AE_{SS}$), Autoencoder-L2 ($AE_{L2}$) \cite{mvtec}, VEVAE \cite{vavae}, CAVGA-R \cite{eatt}, Student \cite{stu}, Superpixel Masking and Inpainting (Superpixel) \cite{super}. As shown in Table 3, the proposed method consistently outperformed all other existing methods evaluated in AUROC. Reconstruction-based methods such as AE and Superpixel used the reconstruction loss as the anomaly score. Using these methods, if the latent space of the training model is large, the reconstruction performance of the anomaly class can be enhanced. In VEVAE and CAVGA, an attention map was obtained using grad-cam \cite{grad}. However, the attention map extracted using grad-cam could not focus on the important region if the model is trained incorrectly. Additionally, since grad-cam-based methods generally generate attention maps through methods that are not based on image segmentation loss, it is difficult to extract an exact important region. Therefore, these methods show lower performance compared to the proposed method. In addition, compared to patch-based methods such as \cite{stu}, which changes the performance according to the patch size, the proposed method achieved higher performance.

\begin{table}
\begin{center}
\label{table:headings}
\renewcommand{\tabcolsep}{4pt}
\makeatletter
\def\hlinewd#1{%
\noalign{\ifnum0=‘}\fi\hrule \@height #1 \futurelet
\reserved@a\@xhline}
\newcommand{\hthickline}{\hlinewd{1pt}}
\newcommand{\hthinline}{\hlinewd{.2pt}}
\makeatother
\newcolumntype{Z}{>{\centering\arraybackslash}X}
\begin{tabularx}{\linewidth}{c||Z Z Z Z}
\hthickline
Scenarios &(a) &(b) &(c) \\
\hline\noalign{\smallskip}
\hline
$L_{Adv}^{ano}$ &\checkmark &-  &\checkmark \\
$L_{Att}$ &- &\checkmark &\checkmark\\
Performance &0.75 &0.87 &0.92 \\\hline
\end{tabularx}
\end{center}
\caption{Performance comparison for the proposed method with various conditions in terms of segmentation AUROC on the MVtec AD dataset (a) with $L_{Adv}^{ano}$, (b) with $L_{Att}$, and (c) with $L_{Adv}^{ano}$ and $L_{Att}$.}
 \vspace{-0.2cm}
\end{table}

Fig. \ref{fig5} shows a qualitative evaluation of the proposed method. Fig. \ref{fig5} (a) shows the normal image, attention map, and GT. Unlike the CIFAR10 dataset, in the MVtec AD dataset, normal samples have a constant image distribution. Therefore, the proposed method easily learned the normal data distribution, and the attention map constantly activated all regions. As shown in Fig. \ref{fig5}. (b), the attention map of the anomaly image activated the normal region and deactivated the useless region. Compared to existing methods, the proposed method had a separate module that directly generates an attention map, and hence, the attention map that is more similar to that of GT was obtained.

In addition, we compared the anomaly detection performance between the proposed and existing methods. As shown in Table 3, the proposed method outperformed the classification accuracy of existing methods by 5.9\%. Existing methods used the maximum reconstruction loss value or the attention map value for the anomaly detection score. Therefore, if the attention map or the reconstructed image is generated incorrectly, the detection performance decreased. In contrast to the existing methods, the proposed method performed segmentation and detection separately on the attention network and ADGAN, respectively. Therefore, unlike existing methods, the proposed ADGAN performed anomaly detection correctly because the input feature distribution did not change significantly even though part of the attention map is generated incorrectly.

\section{Ablation Study}
In order to investigate the effectiveness of each loss in the proposed method, we carried an ablation study. Specifically, four scenarios were considered to verify the effectiveness of the adversarial anomaly loss and the attention loss on the anomaly segmentation performance, (a) with $L_{Adv}^{ano}$, (b) with $L_{Att}$, and (c) with $L_{Adv}^{ano}$ and $L_{Att}$. In addition, to investigate the effectiveness of ADGAN on anomaly detection performance, ADGAN was removed and anomaly detection was performed only with the attention network. The MVtec AD dataset was used to evaluate the segmentation performance, and the CIFAR10 dataset was used to evaluate the detection performance. The effectiveness of each loss was verified in the subsection below.

\begin{figure}[t]
\begin{center}
\includegraphics[width=0.8\linewidth]{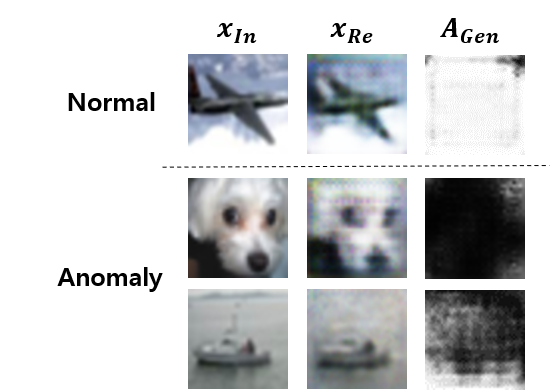}
\end{center}
   \caption{Output images reconstructed from the attention network and the attention map. The normal class of CIFAR10 is airplane.}
   \vspace{-0.3cm}
\label{fig7}
\end{figure}

\subsection{Effectiveness of Adversarial Anomaly Loss and Attention Loss}
In the proposed method, the adversarial anomaly loss $L_{Adv}^{ano}$ and the attention loss $L_{Att}$ were used in the learning process to focus on the normal region. Adversarial anomaly loss was used on the attention network discriminator to deactivate the useless region. In the method with $L_{Att}$ in which $L_{Adv}^{ano}$ was removed, the discriminator was learned using only the normal image and the attention map. The attention map of the normal sample always regarded all regions as active. Therefore, as shown in Fig. \ref{fig6} (b), the attention map of the model in which adversarial anomaly loss was not used can activate the useless region. Additionally, attention loss is used to accurately distinguish between normal and anomaly regions. As shown in Fig. \ref{fig6} (a), the model without attention loss did not accurately distinguish a normal region. However, as shown in Fig. \ref{fig6} (c), the proposed model applying the attention loss accurately distinguished the useless region. Therefore, we confirmed that the additional losses of the proposed method could be applied for the intended purpose, and Table 4 shows the quantitative evaluation of these results. In Table 4, (a) and (b) indicate the performance of the model by removing the attention loss and adversarial anomaly loss, respectively. Table 4 (c) shows that the model to which both losses are applied has a high performance.

\begin{table}
\begin{center}
\label{table:headings}
\renewcommand{\tabcolsep}{4pt}
\makeatletter
\def\hlinewd#1{%
\noalign{\ifnum0=‘}\fi\hrule \@height #1 \futurelet
\reserved@a\@xhline}
\newcommand{\hthickline}{\hlinewd{1pt}}
\newcommand{\hthinline}{\hlinewd{.2pt}}
\makeatother
\newcolumntype{Z}{>{\centering\arraybackslash}X}
\begin{tabularx}{\linewidth}{c||Z Z Z Z}
\hthickline
Scenarios &AUROC \\
\hline\noalign{\smallskip}
\hline
Using reconstruction loss &0.634  \\
Using discriminator output &0.632 \\
Using ADGAN &0.823 \\\hline
\end{tabularx}
\end{center}
\vspace{-0.2cm}
\caption{Performance comparison for the proposed method with various conditions in terms of AUROC on the CIFAR10 dataset.}

\end{table}
\subsection{Effectiveness of ADGAN}
To confirm the effectiveness of ADGAN, anomaly detection was performed without ADGAN, with only the attention network remaining. The models without ADGAN used the discriminator output and the reconstruction loss in the attention network as the anomaly score. As shown in Table 5, the proposed model using ADGAN outperformed other models. The learning process of ADGAN was the same as that of the existing GAN except that the image feature removing the useless region is an input. Therefore, we confirmed that the process of removing a useless region from the image improves the performance of the anomaly detection.

In addition, Fig. \ref{fig7} shows that the reconstruction loss-based model fails to detect the anomaly class. The anomaly reconstructed images of Fig. \ref{fig7} had low reconstruction loss even though input images belong to an anomaly class. Therefore, it was difficult to detect such anomaly images with existing methods. However, the proposed method removed the anomaly class region even in well reconstructed images. This image distribution differed from the distribution learned by ADGAN. Therefore, ADGAN detected the anomaly data more sensitively than existing methods.

\section{Conclusion}
This paper proposed a novel two-stage network consisting of the attention network and ADGAN to remove the useless region unrelated to the normal class and easily detect the anomaly class. In the proposed process, the attention network was learned using the hard augmentation-based anomaly generation to generate an attention map that represents the region with the normal class distribution. Through this attention map, ADGAN learned the normal class distribution after removing the useless region. Moreover, since the attention map of the anomaly class was generated differently from the normal class distribution, anomaly images that are not detected by the existing methods were identified. Experimental results show that the proposed method outperformed the existing SOTA anomaly detection methods on the CIFAR-10 and MVTec AD datasets.

{\small
\bibliographystyle{ieee_fullname}
\bibliography{egpaper_final.bbl}
}

\end{document}